\documentclass[letterpaper, 10 pt, conference]{ieeeconf}
\IEEEoverridecommandlockouts    
\usepackage{hyperref}
\usepackage{threeparttable}
\usepackage{algorithm}
\usepackage{cite}
\hypersetup{
    colorlinks=true,
    linkcolor=magenta,
    filecolor=magenta,      
    urlcolor=magenta,
}

\usepackage{graphics}           
\usepackage{times}              
\usepackage{amsmath}            
\usepackage{amssymb}            
\usepackage{graphicx}
\usepackage{algorithm}
\usepackage[noend]{algpseudocode}
\usepackage{booktabs}
\usepackage{color}
\definecolor{instructioncolor}{rgb}{.5,.5,.5}

\usepackage[font=small]{caption}

\def\secref#1{Sec.~\ref{#1}}
\def\figref#1{Fig.~\ref{#1}}
\def\tabref#1{Tab.~\ref{#1}}
\def\eqref#1{Eq.~(\ref{#1})}
\def\algref#1{Alg.~\ref{#1}}

\makeatletter
\usepackage{xspace}
\DeclareRobustCommand\onedot{\futurelet\@let@token\@onedot}
\def\@onedot{\ifx\@let@token.\else.\null\fi\xspace}

\def\etal{{et al}\onedot}
\makeatother

\usepackage{array}
\newcolumntype{L}[1]{>{\raggedright\let\newline\\\arraybackslash\hspace{0pt}}m{#1}}
\newcolumntype{C}[1]{>{\centering\let\newline\\\arraybackslash\hspace{0pt}}m{#1}}
\newcolumntype{R}[1]{>{\raggedleft\let\newline\\\arraybackslash\hspace{0pt}}m{#1}}

\usepackage{bm}
\usepackage{multirow}
\usepackage{rotating}  

\title{\LARGE \bf Explicit Interaction for Fusion-Based Place Recognition}

\author{Jingyi Xu$^{1}$, Junyi Ma$^{1}$, Qi Wu$^{1}$, Zijie Zhou$^{2}$, Yue Wang$^{3}$, Xieyuanli Chen$^{4}$, and Ling Pei$^{1 *}$ 
  \thanks{This work was supported in part by the National Nature Science Foundation of China (NSFC) under Grant No.62273229 and No.61873163 separately.}
  \thanks{$^{1}$Jingyi Xu, Junyi Ma, Qi Wu, and Ling Pei are with the Shanghai Jiao Tong University.}
  \thanks{$^{2}$Zijie Zhou is with the Beijing Institute of Technology.}
  \thanks{$^{3}$Yue Wang is with the Zhejiang University.}
  \thanks{$^{4}$Xieyuanli Chen is with the National University of Defense Technology.}
  \thanks{$^*$Corresponding author: Ling Pei (ling.pei@sjtu.edu.cn)}
}

\begin{document}
\maketitle

\IEEEpeerreviewmaketitle
\thispagestyle{empty}
\pagestyle{empty}

\begin{abstract}

Fusion-based place recognition is an emerging technique jointly utilizing multi-modal perception data, to recognize previously visited places in GPS-denied scenarios for robots and autonomous vehicles. Recent fusion-based place recognition methods combine multi-modal features in implicit manners. While achieving remarkable results, they do not explicitly consider what the individual modality affords in the fusion system. Therefore, the benefit of multi-modal feature fusion may not be fully explored. In this paper, we propose a novel fusion-based network, dubbed EINet, to achieve \underline{e}xplicit \underline{i}nteraction of the two modalities. EINet uses LiDAR ranges to supervise more robust vision features for long time spans, and simultaneously uses camera RGB data to improve the discrimination of LiDAR point clouds. In addition, we develop a new benchmark for the place recognition task based on the nuScenes dataset. To establish this benchmark for future research with comprehensive comparisons, we introduce both supervised and self-supervised training schemes alongside evaluation protocols. We conduct extensive experiments on the proposed benchmark, and the experimental results show that our EINet exhibits better recognition performance as well as solid generalization ability compared to the state-of-the-art fusion-based place recognition approaches. Our open-source code and benchmark are released at: \url{https://github.com/BIT-XJY/EINet}.
\end{abstract}

\section{Introduction}
\label{sec:intro}

Place recognition is a trendy technique to provide prior locations for SLAM~\cite{chen2019suma++, deng2023nerf, xu2022fast} and global localization~\cite{cao2022end, chen2021range, yin2019mrs} in autonomous navigation systems, especially in the GPS-denied scenes. Vision-based place recognition (VPR)~\cite{arandjelovic2016netvlad, keetha2023anyloc, zhu2023r2former, berton2023eigenplaces, izquierdo2023optimal} has been extensively studied due to the low-cost and simple use of camera sensors. However, camera sensors are vulnerable to environmental factors such as lighting and weather, which can negatively impact descriptor extraction in various environmental changes during place recognition tasks. In contrast, LiDAR-based place recognition (LPR)~\cite{uy2018pointnetvlad, liu2019lpd, chen2021overlapnet, ma2023cvtnet, kong2024sc} overcomes sensitivity to environmental conditions to some degree but lacks the richness of appearance and texture information. As the combination of VPR and LPR, fusion-based place recognition (FPR)~\cite{lu2020pic, komorowski2021minkloc++, lai2022adafusion, liu2022mff, zhou2023lcpr} integrates information from multiple modalities, images and point clouds specifically. They preserve the respective strengths of each modality and thus enhance recognition performance. However, the explicit interaction of different modalities remains undiscovered among the existing fusion frameworks that only operate implicit feature fusion, and a new explainable fusion strategy needs to be further explored for FPR tasks. 


\begin{figure}
  \centering
  \includegraphics[width=1\linewidth]{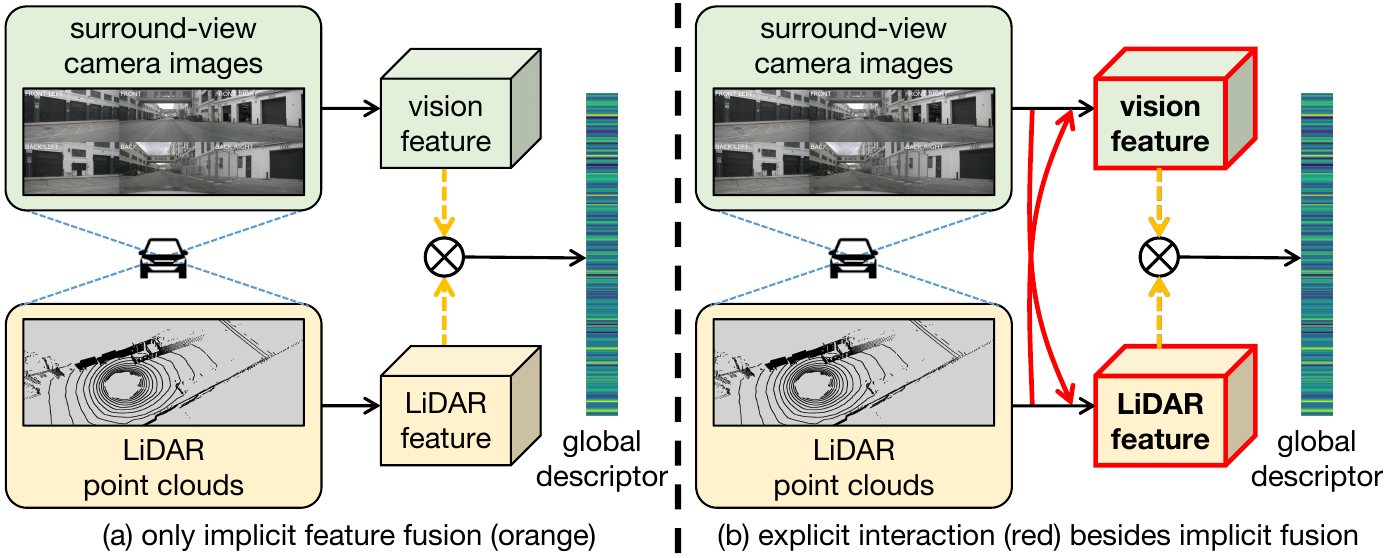}
  \caption{Different from existing implicit fusion methods, our proposed method leverages the strengths of the LiDAR modality (robustness from range) and camera modality (richness from appearance) using explicit interaction for generating place descriptors.}
  \label{fig: motivation}
  \vspace{-0.6cm}
\end{figure}


The two main questions that need to be answered are: \textit{what each modality actually affords positively for possible explicit interaction of place recognition}, and \textit{how to achieve reasonable explicit interaction based on the advantages of the two modalities}.  
It is clear that these two modalities have distinct affordances: LiDAR provides geometric information that is robust over long periods, while cameras offer appearance and texture information that is more distinctive. The key bottleneck lies in the mutual interaction between LiDAR and camera features, and we advocate that an explicit and explainable fusion can further promote place representation compared to implicit fusion methods such as weights adapting \cite{lai2022adafusion} or end-to-end learning \cite{zhou2023lcpr}. To this end, we utilize LiDAR ranges as sparse depth supervision for camera feature extraction, and meanwhile use camera-image-based appearance rendering for colored point cloud generation.
Specifically, a novel explicit interaction network for fusion-based place recognition, dubbed EINet, is proposed to explicitly explore the strengths of both modalities. We integrate two types of interaction, sparse depth supervision and appearance rendering, into a single framework to enhance the descriptive capability of the final global descriptors, as shown in \figref{fig: motivation}. 
In addition, we build a new benchmark, namely NUSC-PR, on the nuScenes dataset \cite{caesar2020nuscenes} for general place recognition tasks, to facilitate the advancements in this emerging domain. NUSC-PR takes into account both supervised and self-supervised paradigms, and also provides the standard evaluation protocol to compare baseline methods.




In sum, the main contributions of our work are threefold:

\begin{itemize}

\item We propose EINet, a novel fusion-based place recognition network with an explicit and explainable interaction mechanism between LiDAR and camera sensors, aiming to jointly leverage the superiority of multi-modal features for generating global descriptors.



\item A new benchmark for FPR called NUSC-PR is proposed to provide two standard training schemes alongside evaluation metrics, which can be used as a foundation tool for future place recognition research.  

\item We conduct comprehensive experiments with detailed analysis to validate that our EINet outperforms the current state-of-the-art methods, including the FPR baselines using implicit feature fusion, and has strong generation ability across different locations.
\end{itemize}

\section{Related Work}
\label{sec:related}

The rapid development of place recognition have been comprehensively documented in surveys over the years \cite{lowry2015visual, yin2023survey, hu2023location}. In this section, we review place recognition from the following three aspects: vision-based, LiDAR-based, and fusion-based methods. Compared to traditional place recognition paradigms (mostly depending on hand-crafted features \cite{arandjelovic2013all, galvez2012bags, he2016m2dp, kim2018scan, cop2018delight}), deep learning-based approaches show promising ability in both recognition accuracy and running efficiency, which we mainly focus on in the three mentioned aspects.

Vision-based place recognition is commonly considered as an image retrieval problem, wherein the database image most similar to the current query image is retrieved. Some prior works \cite{arandjelovic2016netvlad, chen2014convolutional, leyva2021generalized, garg2021your} have laid the foundation for VPR. Especially for basic feature aggregation, NetVLAD by Arandjelovic \etal \cite{arandjelovic2016netvlad} provides a general skill, which is afterward utilized by most descriptor-based methods \cite{khaliq2022multires,ma2022overlaptransformer,liu2019lpd,keetha2023anyloc}. For example, MultiRes-NetVLAD \cite{khaliq2022multires} aggregates a union of features across multiple resolutions using argumentated NetVLAD representation. Recently, more advanced works have been proposed. R2Former proposed by Zhu \etal \cite{zhu2023r2former} uses a novel transformer-based reranking strategy to further improve the recognition accuracy compared to the conventional VPR pipeline. EigenPlaces proposed by Berton \etal \cite{berton2023eigenplaces} extracts descriptors with viewpoint robustness by explicitly clustering the training data. Keetha \etal \cite{keetha2023anyloc} propose AnyLoc, which utilizes pre-trained foundation models to generate descriptors in structured and unstructured environments. Similarly, DINOv2 SALAD by Izquierdo \etal \cite{izquierdo2023optimal} also integrates the foundation model into the place recognition framework, but replaces the commonly used NetVLAD \cite{arandjelovic2016netvlad} with a novel feature aggregation approach based on the optimal transport theory.

Compared to the VPR, LiDAR-based place recognition is robust to environmental changes, such as illumination, weather, and seasons. Uy \etal \cite{uy2018pointnetvlad} propose PointNetVLAD by combining existing PointNet \cite{qi2017pointnet} with NetVLAD to solve the place retrieval problem. LPD-Net by Liu \etal \cite{liu2019lpd} extracts local features, and aggregates them using a graph neural network, to generate global features. OT series including OverlapNet \cite{chen2021overlapnet}, OverlapTransformer \cite{ma2022overlaptransformer}, SeqOT \cite{ma2022seqot}, and CVTNet \cite{ma2023cvtnet} utilize the transformer \cite{vaswani2017attention} to improve recognition performance while maintaining the yaw-angle invariance. BEVPlace by Luo \etal \cite{luo2023bevplace} is also designed to generate rotation-invariant descriptors exploiting bird’s eye view (BEV), while simultaneously exploring the correlation of global feature distances and geometric distances. Cui \etal \cite{cuibow3d} build the bag of words based on the devised 3D LiDAR features for place retrieval. 
Considering texture enhancement in robotics, Xia \etal \cite{xia2023text2loc} newly develop a text-to-place retrieval method that first uses the semantic relationship between laser points and texture description for place recognition.

Recently, cross-modal place recognition \cite{zheng2023i2p, lee20232, yu2020monocular, yin2021i3dloc} has aroused great interest, unifying the cross-modal data into the same modality. 
While producing remarkable results, directly fusing multi-modal inputs to recognize places rather than modality unifying for query and references still exhibits the most superior performance \cite{zhou2023lcpr}. One of the early fusion-based approaches,
PIC-Net proposed by Lu \etal \cite{lu2020pic} integrates the global channel attention to directly fuse the features of camera images and LiDAR point clouds. Pan \etal \cite{pan2021coral} propose CORAL-VLAD, a network fusing the structural features and vision features in the consistent BEV for place recognition. MinkLoc++ by Komorowski \etal \cite{komorowski2021minkloc++} introduces a late fusion approach to process each modality separately and fuse them in the final part. AdaFusion by Lai \etal \cite{lai2022adafusion} utilizes multi-scale attention to learn the fusion weights between camera and LiDAR modalities in different environments. Compared to AdaFusion, MFF-PR proposed by Liu \etal \cite{liu2022mff} introduces more comprehensive features including semantic, instance, topological, and image texture features, to further improve the robustness and scene expression abilities for place recognition. More recently, Zhou \etal \cite{zhou2023lcpr} introduce a novel multi-scale attention-based fusion framework LCPR, to generate discriminative and approximately yaw-rotation invariant global descriptors for place recognition. 

To pursue good place recognition performance, our method also adopts the fusion-based stream to generate place descriptors. Different from the above-mentioned FPR methods which advocate implicit integration of multi-modal features, our method instead achieves explicit interaction based on the explainable advantage complementary of multi-modal sensor data.


\section{Method}
\label{sec:method}

\begin{figure*}
  \centering
  \includegraphics[width=0.9\linewidth]{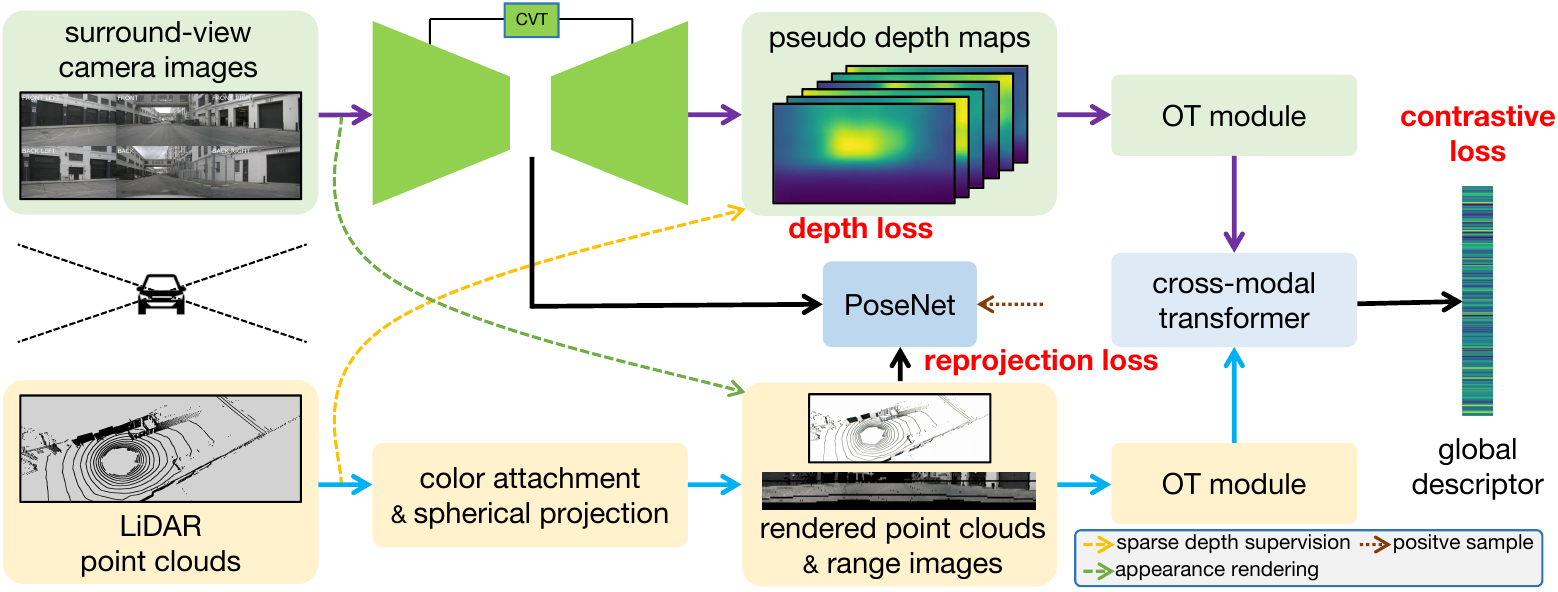}
  \caption{Pipeline overview of our proposed EINet. The purple and blue arrows represent the camera branches and the LiDAR branches respectively. The camera sensors provide RGB information for rendered point clouds and range images in the LiDAR branch (dotted green arrow), while the LiDAR sensor helps to supervise pseudo depth in the camera branch (dotted orange arrow), achieving explicit interaction in the fusion-based place recognition framework.}
  \label{fig:aianet}
  \vspace{-0.6cm}
\end{figure*}

In this section, we first present the overall system of our proposed place recognition approach. Then, we investigate the explicit form of LiDAR and camera interaction in the fusion system. Lastly, we introduce the multiple loss functions used to train EINet.

\subsection{EINet Architecture}
\label{sec:AIANet}

The overall architecture of our proposed EINet is illustrated in \figref{fig:aianet}. EINet uses surround-view camera images and LiDAR point clouds as joint inputs to generate final global descriptors for place retrieval. The camera branch first transforms the multiple input images to pseudo depth maps based on an encoder-decoder structure, under the sparse depth supervision from LiDAR point clouds. Following SurroundDepth \cite{wei2023surrounddepth}, we add cross-view transformer (CVT) blocks between the encoder and decoder, to enhance the correlation between surround-view images. The multiple pseudo depth maps are then transformed into a holistic range image in the LiDAR coordinate system for the following compression and feature fusion.
The LiDAR branch first combines the RGB information provided by cameras and spherical projection to produce rendered point clouds as well as range images. The rendered range images from LiDAR and the counterpart transformed from pseudo depth are both compressed by OT modules (OverlapTransformer \cite{ma2022overlaptransformer} without Global Descriptor Generator) in the respective branch. Then the two sentence-like features are absorbed by the cross-modal transformer, which contains stacked cross transformer blocks and NetVLAD with MLP \cite{ma2023cvtnet}, to produce the descriptor vector for place recognition. Notably, we further utilize an additional PoseNet to explicitly capture the relative pose between the query and the positive samples. More details can be found in \secref{sec:Loss}.



\subsection{Explicit Interaction}
\label{sec:Affords}

Explicit interaction lies in the complementary advantages of the two modalities. LiDAR offers more stable geometric features that can be used to improve the robustness of camera features in long-time gaps. Cameras offer rich appearance and texture information, which can conversely be used to enhance the distinctiveness of point cloud features. Based on the above analysis, we first specify what LiDAR and camera explicitly afford to each other in the cross-modal interaction of our fusion-based place recognition framework.

\begin{figure}
  \centering
  \includegraphics[width=1\linewidth]{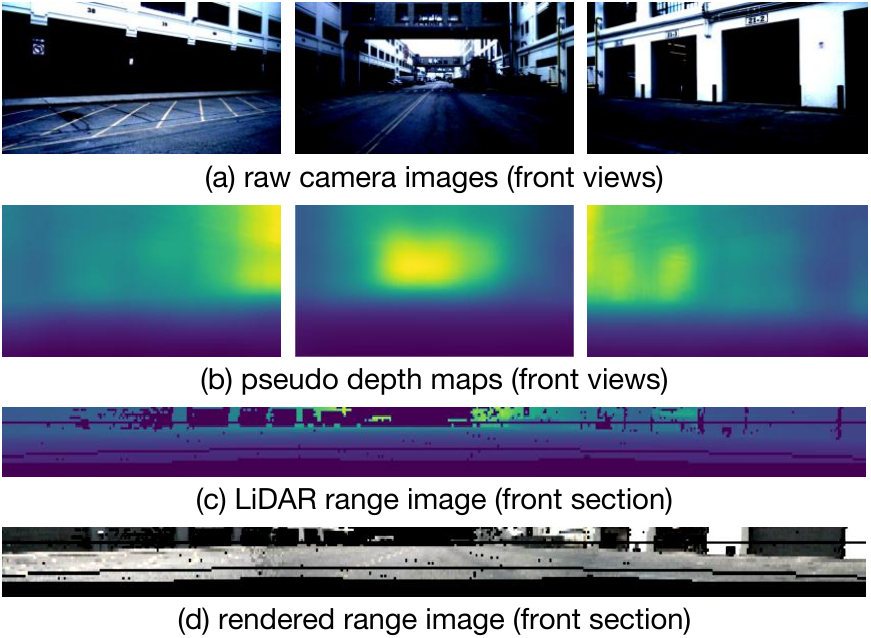}
  \caption{An example of the raw camera images, pseudo depth maps, raw LiDAR range image, and rendered range image in explicit interaction.}
  \label{fig:affordance}
  \vspace{-0.8cm}
\end{figure}

\textbf{Sparse depth supervision.} We first utilize ranges captured by LiDAR as sparse depth supervision ($\mathcal{I}\oplus\mathcal{P}\rightarrow\mathcal{D}$) for the camera feature extraction.
As shown in \figref{fig:aianet}, LiDAR explicitly helps to guide the image encoder and decoder in the camera branch to output pseudo depth maps $\mathcal{D}$ from the input image $\mathcal{I}$. We achieve this by projecting raw LiDAR point clouds $\mathcal{P}$ to the camera image planes to supervise depth generation with depth loss sparsely (see \secref{sec:Loss}). We release the dense constraints proposed by the previous depth estimation works \cite{wei2023surrounddepth, zhou2017unsupervised} such as photometric reprojection loss, and thus only generate pseudo depth maps instead of common depth maps by strict definition (see \figref{fig:affordance}).
This helps to keep mixed-state features in the camera branch that are geometry-aware as well as appearance-aware to balance robustness and distinctiveness. Notably, the predicted depth maps are scale-aware since we use real distance observed by LiDAR for supervision.

\textbf{Appearance rendering.} Conversely, we propose appearance rendering ($\mathcal{P}\oplus\mathcal{I}\rightarrow\mathcal{P}_{color}$) as what cameras explicitly provide to LiDAR feature generation. Although geometric features directly extracted from point clouds are less affected by environmental disturbance, they are naturally sparse and lack observation distinctiveness. Therefore, we use camera images to attach RGB channels to point clouds $\mathcal{P}$, leading to colored point clouds $\mathcal{P}_{color}$, which are then transformed to rendered range images by spherical projection (see \figref{fig:affordance}). Color attachment for point clouds can be implemented by the similar operation in sparse depth supervision. We only need to project raw laser points to the camera image planes and sample RGB values respectively. The projection function is the same for different branches, which helps save computing resources in our proposed framework.

The above-mentioned interaction including sparse depth supervision and appearance rendering are the foundation to explicitly complement each other's advantages of the two modalities. We further jointly use them in the unified EINet to achieve explainable fusion-based place recognition. 
The image features and the point cloud features are tightly coupled by the interaction and then fused by the cross-modal transformer, leading to global descriptors suitable for place retrieval with different time gaps. 
In the proposed framework, we do not directly fuse vanilla camera and LiDAR features compared to the previous works \cite{zhou2023lcpr,lai2022adafusion}, but let the LiDAR branch assist the camera branch in generating geometry-aware features which are then fused back to the LiDAR features. Similarly, we also let the camera branch assist the LiDAR branch in producing appearance-aware features, which are then fused with the features in the camera branch. The explicit interaction therefore helps the following transformer to extract correlations and find consistency across modalities.
The effectiveness and validity of our proposed explicit interaction are further guaranteed by the setups of training losses, which will be introduced in \secref{sec:Loss}.

\subsection{Loss Functions}
\label{sec:Loss}
We use three types of losses to train EINet, harnessing the positive effects of explicit interaction.

\textbf{Depth loss.} For the input point cloud $\mathcal{P}$, a laser point $p_i \in \mathcal{P}~(i \in [1,N_{\text{pc}}])$ is first projected to the camera coordinate system by extrinsic parameters, leading to $p_i^{\text{c}}=(x_i^{\text{c}},y_i^{\text{c}},z_i^{\text{c}})$ with depth $d_i=z_i^{\text{c}}$. Then $p_i^{\text{c}}$ is transformed to the pixel $(u_i, v_i)$ in the corresponding image $\mathcal{I}_j~(j \in [1,N_{\text{cam}}])$ by the camera intrinsic parameters. Then we calculate the depth loss using $L1$ distance by
\begin{align}
  \mathcal{L}_{\text{d}} = \sum_{i,j}\mathbb{I}(|d_i-\mathcal{D}_j(u_i, v_i)|), \label{eq:range_loss}
\end{align}
where $\mathcal{D}_j$ is the pseudo depth map of the $j$th camera. $\mathbb{I}(a)=a$ if $(u_i, v_i) \in \mathcal{I}_j$ and $\mathbb{I}(\cdot)=0$ otherwise. The depth loss concretely represents how LiDAR ranges function on the camera branch in our proposed EINet.

\textbf{Contrastive loss.} We choose commonly used contrastive loss \cite{ma2022overlaptransformer, ma2023cvtnet, liu2019lpd} to supervise global descriptor generation. For each training tuple, we utilize one query descriptor $\mathcal{G}_\text{q}$, $n_\text{pos}$ positive descriptors {$\mathcal{G}_\text{p}$}, and $n_\text{neg}$ negative descriptors $\mathcal{G}_\text{n}$ to calculate lazy triplet loss by
\begin{align}
  \mathcal{L}_\text{t} = n_\text{pos}(\alpha+\text{max}(\text{dis}(\mathcal{G}_\text{q}, \mathcal{G}_\text{p})))-\sum_{n_\text{neg}}\text{dis}(\mathcal{G}_\text{q}, \mathcal{G}_\text{n}), \label{eq:contrastive_loss}
\end{align}
where $\alpha$ is the margin and $\text{dis}(\cdot)$ is squared Euclidean distance to measure similarity of descriptors. Using the triplet loss we can narrow the gap between feature representations for close-place pairs and enlarge the counterparts for distant-place pairs. In \secref{sec:benchmark}, we will introduce different approaches to select positive and negative samples for the query one in our proposed benchmark.

\textbf{Reprojection loss.} We notice that most existing place recognition methods only use the straightforward way, i.e., the abovementioned contrastive loss, to coarsely capture the relationship of similar places. To make the network explicitly aware of the quantized relation between the query and positive samples, we propose an additional reprojection loss based on the devised PoseNet (as shown in \figref{fig:aianet}). The PoseNet uses the encoded image features of the query sample and a randomly selected positive sample to predict relative poses of the robot $T_\text{e}$ between the two frames. Then $T_\text{e}$ is transformed to the relative LiDAR pose $T_\text{L}$ by extrinsic parameters to compute the reprojection loss by
\begin{align}
  \mathcal{L}_\text{r} = |\mathbb{S}(T_\text{L}\mathcal{P_\text{p}})-\mathbb{S}(\mathcal{P_\text{q}})|,
  \label{eq:reprojection_loss}
\end{align}
where $\mathbb{S}$ represents spherical projection for point clouds to generate range images. We reduce the difference between the query range image and the positive range image reprojected to the query frame to let the network receive supervision signals related to pose quantification. Note that we substantially use the reprojection loss to optimize the image backbone for point cloud alignment, further enhancing the correlation between the two modalities.

The total loss used to train the EINet is the weighted summation of the above three losses, which is computed by
\begin{align}
  \mathcal{L}_\text{all} = \lambda_\text{d} \mathcal{L}_{\text{d}}+ \lambda_\text{t} \mathcal{L}_{\text{t}}+ \lambda_\text{r} \mathcal{L}_{\text{r}},
  \label{eq:weighted_loss}
\end{align}
where $\lambda_\text{d}$, $\lambda_\text{t}$, and $\lambda_\text{r}$ are the weighting parameters.

\section{The NUSC-PR Benchmark}
\label{sec:benchmark}

Although place recognition approaches have been widely proposed in recent years, a unified benchmark is lacking to standardize both the supervised and self-supervised learning schemes and evaluation protocols. To this end, we further build a new benchmark, NUSC-PR, based on the publicly available nuScenes dataset \cite{caesar2020nuscenes}, with the standardized dataset format and evaluation protocol. 

\textbf{Data organization.} We first re-organize the original nuScenes dataset to a new format, to build the training set and the test set. Regarding learning-based place recognition methods, the supervised learning scheme is widely exploited based on manually sampling positive and negative places with distance metrics offline~\cite{zhou2023lcpr, cai2022autoplace}. It needs ground-truth positions (always provided by GPS-based positioning systems) to quantify the similarity between the query and reference samples. Our proposed NUSC-PR benchmark continues adopting this supervised learning scheme, for which the data organization is shown in \algref{alg:pipline1}.
In addition to the commonly used supervised manner with distance metrics, we also notice that almost all learning-based place recognition models can be trained online using a self-supervised manner, where the positive and negative samples are split by time metrics automatically. In this case, the collection time determines whether a sample is positive or negative, and the time period strictly corresponds to the number of samples since the collection frequency is fixed over the dataset. If the time interval between the query sample and the reference counterpart is less than the preset threshold, the reference sample is regarded as positive for its query place. Therefore, the training process can be implemented online even when the autonomous vehicle is driving in a GPS-denied environment since this paradigm does not need high-quality ground-truth positions. 
However, we still need the ground-truth positions of all the samples to conduct the test set for offline evaluation. 
The overall data organization for the self-supervised learning scheme is shown in \algref{alg:pipline2}.
To summarize, NUSC-PR contains two types of data organization for supervised and self-supervised learning schemes. In \secref{sec:experiments}, we conduct comprehensive experiments based on the datasets built by both \algref{alg:pipline1} and \algref{alg:pipline2}. 

\begin{algorithm}[t]
\small
\caption{Data generation pipeline for the~\textbf{supervised} learning scheme in the NUSC-PR benchmark.}
\hspace*{0.02in} {\bf Input:}
Driving scenes $\{\mathcal{S}\}_{i=0}^{N_s}$, data samples $\{s\}_{j=0}^{N_{\mathcal{S}_i}} \in \mathcal{S}_i$, distance interval $\delta$, date threshold $\gamma$, positive distance threshold $\rho_{\text{pos}}$, negative distance threshold $\rho_{\text{neg}}$, the numbers of positive and negative samples $n_\text{pos}$ and $n_\text{neg}$ for each query;\\
\hspace*{0.02in} {\bf Output:}
The training set $\mathcal{Q}$ and test set $\mathcal{K}$ for supervised learning;
\begin{algorithmic}[1]
\State Initialize the database set $\mathcal{A}=\phi$, the training query set $\mathcal{B}=\phi$, the test query set $\mathcal{C}=\phi$.
\For{each scene $\mathcal{S}_i \in \{\mathcal{S}\}_{i=0}^{N_s}$}
    \For{each data sample $s_j \in \{s\}_{j=0}^{N_{\mathcal{S}_i}}$}
    \State $\mu = \text{min}(s_j, \mathcal{A})$;
       \If{$\mu \geq \delta$ or $\mathcal{A}=\phi$}
        \State Append $s_j$ to the database set $\mathcal{A}$;
       \ElsIf{the date of $s_j$ is before $\gamma$}
        \State Append $s_j$ to the training query set $\mathcal{B}$;
       \Else
        \State Append $s_j$ to the test query set $\mathcal{C}$;
        \EndIf
    \EndFor
\EndFor
\State Initialize the training set $\mathcal{Q}=\phi$;
\For{each train query sample $a_i \in \mathcal{A}$}
    \State Initialize the positive set $\mathcal{V}_i^{\text{pos}}=\phi$ and the negative set $\mathcal{V}_i^{\text{neg}}=\phi$ for the current query;
    \State Use K-Nearest Neighbours (KNN) to find the potential positive samples with $\rho_{\text{pos}}$, which are randomly selected to $\mathcal{V}_i^{\text{pos}}$ with the number of $n_\text{pos}$;
    \State Find the potential negative samples within the complementary set of KNN results using $\rho_{\text{neg}}$, which are randomly selected to $\mathcal{V}_i^{\text{neg}}$ with the number of $n_\text{neg}$;
    \State Append $\{i:[\mathcal{V}_i^{\text{pos}},\mathcal{V}_i^{\text{neg}}]\}$ to the training set $\mathcal{Q}$;
\EndFor
\State Initialize the test set $\mathcal{K}=\phi$;
\For{each test query sample $b_i \in \mathcal{B}$}
    \State Initialize the ground-truth set $\mathcal{V}_i^{\text{gt}}=\phi$;
    \State Use KNN to find the ground-truth samples $\mathcal{V}_i^{\text{gt}}$ with $\rho_{\text{pos}}$;
    \State Append $\{i:[\mathcal{V}_i^{\text{gt}}]\}$ to the test set $\mathcal{K}$;
\EndFor
\State Optional: randomly split validation set out of test set with a preset proportion.
\State \Return $\mathcal{Q}$ and $\mathcal{K}$.
\end{algorithmic}
\label{alg:pipline1}
\end{algorithm}

\begin{algorithm}[t]
\small
\caption{Data generation pipeline for the \textbf{self-supervised} learning scheme in the NUSC-PR benchmark.}
\hspace*{0.02in} {\bf Input:}
Driving scenes $\{\mathcal{S}\}_{i=0}^{N_s}$, data samples $\{s\}_{j=0}^{N_{\mathcal{S}_i}} \in \mathcal{S}_i$, date threshold $\gamma$, positive distance threshold $\rho_\text{pos}$, negative time threshold $\sigma_\text{neg}$, the numbers of positive and negative samples $n_\text{pos}$ and $n_\text{neg}$ for each query;\\
\hspace*{0.02in} {\bf Output:}
The training set $\mathcal{Q}$ and test set $\mathcal{K}$ for self-supervised learning;
\begin{algorithmic}[1]
\State Initialize the database set $\mathcal{A}=\phi$, the training query set $\mathcal{B}=\phi$, the test query set $\mathcal{C}=\phi$, the old scene set $\mathcal{E}_\text{old}=\phi$, the new scene set $\mathcal{E}_\text{new}=\phi$.
\For{each scene $\mathcal{S}_i \in \{\mathcal{S}\}_{i=0}^{N_s}$}
   \If{the date of $\mathcal{S}_i$ is before $\gamma$}
    \State Append $\mathcal{S}_i$ to the old scene set $\mathcal{E}_\text{old}$;
   \Else
    \State Append $\mathcal{S}_i$ to the new scene set $\mathcal{E}_\text{new}$;
    \EndIf
\EndFor
\State Initialize the whole old sample set $\mathcal{V}_{*}^{\text{old}}=\phi$;
\For{each scene $\mathcal{S}_i \in \mathcal{E}_\text{old}$}
    \State Initialize the potential negative set $\mathcal{V}_{*}^{\text{neg}}=\phi$;
    \For{each data sample $s_j \in \{s\}_{j=0}^{N_{\mathcal{S}_i}}$}
        \State Append $s_j$ to $\mathcal{V}_{*}^{\text{old}}$;
        \If{$j \geq \sigma_\text{neg}+n_\text{pos}+n_\text{neg}$}
            \State Initialize the positive set $\mathcal{V}_j^{\text{pos}}=\phi$ and the negative set $\mathcal{V}_j^{\text{neg}}=\phi$ for the current query;
            \State Select $\{s\}_{k=j-n_\text{pos}}^{j-1}$ as $\mathcal{V}_j^{\text{pos}}$;
            \State Randomly select $n_\text{neg}$ samples in $\mathcal{V}_{*}^{\text{neg}}$ as $\mathcal{V}_j^{\text{neg}}$;
            \State Append $s_{j-\sigma_\text{neg}}$ to $\mathcal{V}_{*}^{\text{neg}}$;
            \State Append $\{j:[\mathcal{V}_j^{\text{pos}},\mathcal{V}_j^{\text{neg}}]\}$ to the training set $\mathcal{Q}$;
        \Else
            \State Append $s_j$ to $\mathcal{V}_{*}^{\text{neg}}$;
        \EndIf
    \EndFor
\EndFor
\State Initialize the test set $\mathcal{K}=\phi$;
\For{each scene $\mathcal{S}_i \in \mathcal{E}_\text{new}$}
    \For{each data sample $s_j \in \{s\}_{j=0}^{N_{\mathcal{S}_i}}$}
        \State Initialize the ground-truth set $\mathcal{V}_{j}^{\text{gt}}=\phi$;
        \State Use KNN to find the ground-truth samples $\mathcal{V}_j^{\text{gt}}$ with $\rho_{\text{pos}}$;
        \State Append $\{j:[\mathcal{V}_j^{\text{gt}}]\}$ to the test set $\mathcal{K}$;
    \EndFor
\EndFor
\State Optional: randomly split validation set out of test set with a preset proportion.
\State \Return $\mathcal{Q}$ and $\mathcal{K}$.
\end{algorithmic}
\label{alg:pipline2}
\end{algorithm}

\textbf{Evaluation metrics.}
The average recall rate is a widely used metric to represent the proportion of successful retrieval in place recognition~\cite{zhu2023r2former, izquierdo2023optimal, ma2023cvtnet}. In the NUSC-PR benchmark, we also use the average recall rates (e.g., AR@1, AR@5, AR@10, and AR@20) to evaluate different methods on the test set, which are calculated by
\begin{align}
    \text{AR@}x = \frac{n_\text{suc}}{N_\text{query}} \times 100\%
    \label{eq:AR}
\end{align}
where $N_\text{query}$ is the number of test query samples, and $n_\text{suc}$ is the number of successful recall. Only when there is at least one correct reference sample in the top $x$ retrievals, one successful recall is collected for the numerator of $\text{AR@}x$. We use different test queries and databases to evaluate the supervised learning and self-supervised learning schemes generated by \algref{alg:pipline1} and \algref{alg:pipline2}, leading to respective evaluation protocols.


\section{Experiments}
\label{sec:experiments}

The experiments are conducted to validate the claims that our approach is able to: (i) accurately retrieve previously visited places with large time gaps in driving environments, (ii) exploit explicit interaction of LiDAR and camera modalities to improve the recognition accuracy, (iii) generalize well into the environments of different locations without fine-tunning, and (iv) be implemented online with efficient inference operation.

\subsection{Experimental Setups}
\label{sec:setups}

\textbf{Benchmark details.} We first provide the detailed configurations of \algref{alg:pipline1} and \algref{alg:pipline2} in our proposed NUSC-PR benchmark. We build datasets for the individual locations of different cities, including Boston-Seaport (BS), Singapore-Onenorth (SON), Singapore-Queenstown (SQ), and Singapore-Hollandvillagee (SHV) respectively. We set the date threshold $\gamma$ as 105 days following \cite{zhou2023lcpr, cai2022autoplace}, the positive distance threshold $\rho_{\text{pos}}$ as 9\,m, the numbers of positive and negative samples $n_\text{pos}$ and $n_\text{neg}$ as 2 and 4 respectively. We let the distance interval $\delta = 1$\,m and the negative distance threshold $\rho_{\text{neg}}=18$\,m in the supervised learning scheme, the negative time threshold $\sigma_\text{neg} = 6$ (3\,s) in the self-supervised learning scheme. The detailed configurations of the data organization in the NUSC-PR for different locations are also posted in our open-source code.

\textbf{Baseline setups.} We select the existing camera-based methods including NetVLAD \cite{arandjelovic2016netvlad} and MultiRes-NetVLAD (mrNVLAD) \cite{khaliq2022multires}, LiDAR-based methods including OverlapTransformer (OT) \cite{ma2022overlaptransformer} and CVTNet \cite{ma2023cvtnet}, and fusion-based methods including AdaFusion \cite{lai2022adafusion} and LCPR \cite{zhou2023lcpr} as baselines. We use the default architecture and training parameters of the baseline approaches suggested in their original papers and open sources. We replace the OverlapNetLeg in OT \cite{ma2022overlaptransformer} with the point cloud encoder proposed in \cite{ma2023cvtnet} to adapt 32-beam LiDAR in the nuScenes dataset.

\textbf{EINet setups.} In the camera branch of EINet, we use the resized images with a resolution of 640$\times$352. Following SurroundDepth \cite{wei2023surrounddepth}, we set 4 scales in the image encoder and decoder. We use 4 CVT blocks for each scale between the encoder and decoder, instead of 8 blocks to improve inference efficiency. The pseudo depth maps have the same resolution as the input images. The range image from the multiple pseudo depth maps is 352$\times$1056, fed to the following devised OT module. 
In the LiDAR branch, we project rendered point cloud to the range image with the size of 32$\times$1056, fed to the following OT module with a different structure from the above-mentioned counterpart due to specific input sizes. 
The cross-model transformer module has 2 stacked cross transformers followed by NetVLAD and one MLP. The global descriptor generated by EINet is a 1$\times$256 vector. The PoseNet for auxiliary training has 3 convolution layers that receive the 6$\times$512$\times$11$\times$20 encoded image features of both a query and a randomly selected positive sample, to produce the 6-DOF relative pose between them. Notably, we discard the PoseNet in the test process since it does not contribute to the final descriptor. The loss weights $\lambda_\text{d}$, $\lambda_\text{t}$, and $\lambda_\text{r}$ are set to $0.01$, $1.00$, and $0.01$ respectively. The margin $\alpha$ in $\mathcal{L}_\text{t}$ is set to $0.5$.

All baselines and our proposed methods are trained with a batch size of 7 on 2 NVIDIA A100 GPUs for 20 epochs. We use the ADAM optimizer to optimize our EINet with an initial learning rate of 1e-5 and weight decay of 0.5 applied every 5 steps. More details about training the baselines, as well as EINet can be found in our open-source repository.

\subsection{Assessment on NUSC-PR}
\label{sec:Assessment}

The first experiment supports our claim that EINet can accurately retrieve previously visited places with long time spans in large-scale environments. We compare EINet with the baselines in the proposed NUSC-PR benchmark where the largest time gap between test query with the database is over 120$\,$days. 

\textbf{Evaluation on supervised learning scheme.} The experimental results are shown in \tabref{tab:supervised}, which indicate that our proposed EINet outperforms all the baseline methods using positive and negative samples selected by distance metrics to supervise. Notably, EINet with explicit cross-modal interaction improves the FPR baselines AdaFusion and LCPR using only implicit feature fusion by 10.50\% and 5.77\% in AR@1.
Besides, fusion-based approaches in general have better recognition performance than uni-modal baselines, and LiDAR-based baselines outperform the camera-based counterparts. This demonstrates that fusing multi-model features helps to improve the distinctiveness of global descriptors for place retrieval, and LiDAR can offer stable features with better place description ability than cameras.

\begin{table}[t]
  \centering
  \begin{center}
  	\setlength{\tabcolsep}{1.6mm}
  	\renewcommand\arraystretch{1.1}
    \caption{Evaluation of place recognition performance in the supervised learning scheme on the BS split}
    \vspace{-0.1cm}
    \footnotesize{
        \begin{tabular}{lccccc}
\toprule
\multirow{2}{*}{Methods} & \multirow{2}{*}{Modality$^1$} & \multicolumn{4}{c}{BS split} \\ \cline{3-6} 
                         &                           & AR@1  & AR@5 & AR@10 & AR@20 \\ \midrule
NetVLAD \cite{arandjelovic2016netvlad}                  & C                         & 73.98      & 82.33     & 84.89      & 85.97      \\
mrNVLAD \cite{khaliq2022multires}                 & C                         & 75.31      & 84.49     & 86.40      & 88.23      \\\hline
OT \cite{ma2022overlaptransformer}      & L                         & 74.67      & 84.23     & 87.12      & 89.70      \\
CVTNet \cite{ma2023cvtnet}                  & L                         & 79.20      & 87.96     & 90.35      & 92.44      \\\hline
AdaFusion \cite{lai2022adafusion}               & C+L                       & 80.90      & 87.74     & 90.08      & 92.31      \\
LCPR \cite{zhou2023lcpr}                    & C+L                       & 85.63      & 92.21     & 94.55      & 95.75      \\
EINet (ours)                  & C+L                       & \textbf{91.40}      & \textbf{97.30}     & \textbf{98.75}      & \textbf{99.39}      \\ \bottomrule
\multicolumn{5}{l}{$^1$ C: Camera-only, L: LiDAR-only, C+L: Camera and LiDAR.}
\end{tabular}}

    \label{tab:supervised}
    \end{center}
    \vspace{-0.8cm}
\end{table}

\begin{figure}
\vspace{0.3cm}
  \centering
  \includegraphics[width=1\linewidth]{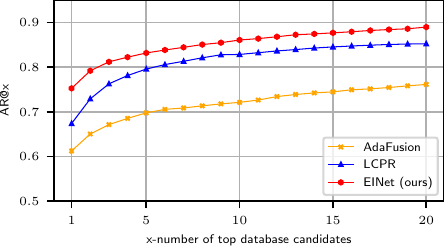}
  \caption{Evaluation of place recognition performance in the self-supervised learning scheme on the BS split.}
  \label{fig:selfsupervised}
  \vspace{-0.8cm}
\end{figure}

\textbf{Evaluation on self-supervised learning scheme.} We compare the fusion-based baselines with EINet in the self-supervised learning scheme of the NUSC-PR. The $\text{AR@}x$ is presented in \figref{fig:selfsupervised}. The recognition performance of all the methods decreases compared with the counterparts posted in \tabref{tab:supervised}, because time metrics provide coarser discernment between positive and negative samples than distance metrics. Our EINet still performs better than the baselines over different recall rates, demonstrating that EINet is more suitable for online self-supervised learning using time metrics.

\subsection{Ablation Study on Explicit Interaction}
\label{sec:ablation}

The experiment supports our claim that our proposed EINet can exploit explicit cross-modal interaction including sparse depth supervision and appearance rendering to enhance recognition performance. In \tabref{tab:ablation}, we compare the holistic EINet with the baseline removing LiDAR-based depth supervision (EINet-rL), and the one removing camera-based appearance rendering (EINet-rC). As can be seen, without the feature enhancement from explicit LiDAR ranges, the performance of EINet-rL significantly drops by 6.45\% on AR@1 compared to EINet. In contrast, EINet-rC still performs worse than the holistic one but outperforms EINet-rL, which indicates that the interaction of sparse depth supervision yields greater advantages than appearance rendering for fusion-based place recognition. The reason could be geometric information captured by LiDAR is more robust for place retrieval in large-scale environments with long time spans than appearance and texture information from camera images. Note that although we remove any type of interaction in this experiment, we retain the fusion-based framework that helps EINet-rL and EINet-rC achieve better performance than the uni-modal baselines in \tabref{tab:supervised}.

\begin{table}[t]
  \centering
  \begin{center}
  	\setlength{\tabcolsep}{3.9mm}
  	\renewcommand\arraystretch{1.2}
    \caption{Ablation study on the explicit interaction}
  \vspace{-0.1cm}
    \footnotesize{
        \begin{tabular}{lcccc}
    \toprule
    \multirow{2}{*}{Methods} & \multicolumn{4}{c}{BS split} \\ \cline{2-5} 
                         & AR@1  & AR@5 & AR@10 & AR@20 \\ \midrule
EINet-rL                 & 84.95      & 93.58     & 95.92      & 98.02      \\
EINet-rC                 & 87.29      & 95.76     & 97.54      & 98.83      \\
EINet                   & \textbf{91.40}      & \textbf{97.30}     & \textbf{98.75}      & \textbf{99.39}      \\ \bottomrule
    \end{tabular}}
    \label{tab:ablation}
  \end{center}
    \vspace{-0.5cm}
\end{table}

\subsection{Study on Generalization Ability}
\label{sec:generalization}

We further implement zero-shot transfer to support the third claim that our EINet has a solid generalization ability across different locations without fine-tuning. We use the supervised scheme in this experiment because the self-supervised scheme can help the place recognition model adaptively fit to new environments through incremental or lifelong learning \cite{cui2023ccl, knights2022incloud} rather than transfer learning. We train all the fusion-based approaches on the BS split and directly evaluate them on the test sets of other splits from SON, SQ, and SHV locations. The results on AR@1 shown in \tabref{tab:locations} demonstrate that EINet generalizes best into unseen locations even in different countries (US $\rightarrow$ SGP). AdaFusion has the worst generalization ability because the weights learned in one location to balance modalities are hard to ensure rationality regarding other locations. Note that there are counterintuitive results on the unseen SON split where the recognition accuracy is higher than the counterparts in the previously seen BS split, which is because the SON split provides easier query-reference pairs in the test set.

\begin{table}[t]
  \centering
  \begin{center}
  	\setlength{\tabcolsep}{4.3mm}
  	\renewcommand\arraystretch{1.2}
    \caption{Comparison of the generalization ability of the fusion-based approaches}
  \vspace{-0.1cm}
    \footnotesize{
        \begin{tabular}{lcccc}
\toprule
\multirow{2}{*}{Methods} & \multicolumn{4}{c}{Locations} \\ \cline{2-5} 
                         & BS    & SON    & SQ   & SHV   \\ \midrule
AdaFusion                & 80.90      & 82.24       & 60.04     & 62.14      \\
LCPR                     & 85.63     & 93.76       & 67.53     & 69.29      \\
EINet (ours)            & \textbf{91.40}      & \textbf{98.29}       & \textbf{73.84}     & \textbf{85.25}     \\ \bottomrule
\end{tabular}}
    \label{tab:locations}
  \end{center}
    \vspace{-0.8cm}
\end{table}

\subsection{Study on Running Efficiency}
\label{sec:runtime}

In this experiment, we provide the inference time of each module in our proposed EINet implemented with Python. The hardware has been explained in \secref{sec:setups}. We calculate the average runtime of the image encoder-decoder, the OT module in the camera branch (OT-C), the OT module in the LiDAR branch (OT-L), the cross-modal transformer (CMT) on the samples of the holistic nuScenes dataset. The results are shown in \tabref{tab:time}. As can be seen, the image decoder is the slowest part among the five modules, which costs 58.23\,ms due to the utilization of CVT blocks. The summation of the other four modules is only 12.63\,ms. Besides, we also calculate the average time (114.70\,ms) of the transformation from the pseudo depth maps to the input of the OT module in the camera branch. Note that we do not consider the spherical projection time in the LiDAR branch because some existing LiDAR sensors can directly deliver raw range images \cite{ma2022overlaptransformer}. Therefore, the total inference time of EINet is around 186.48\,ms (5.36\,Hz), which is efficient for online global localization in real-world applications.

\begin{table}[t]
  \centering
  \begin{center}
  	\setlength{\tabcolsep}{3.0mm}
  	\renewcommand\arraystretch{1.2}
    \caption{Inference time of each module in our proposed EINet}
  \vspace{-0.1cm}
    \footnotesize{
        \begin{tabular}{cccccl}
\toprule
Module & \begin{tabular}[c]{@{}c@{}}Image\\Encoder\end{tabular} & \begin{tabular}[c]{@{}c@{}}Image\\Decoder\end{tabular} & OT-C & OT-L & CMT \\ \midrule
Time [ms]   & 5.14       & 58.23       & 2.14       & 1.39       & 3.96       \\ \bottomrule
\end{tabular}}
    \label{tab:time}
  \end{center}
    \vspace{-0.8cm}
\end{table}


\section{Conclusion}

In this paper, we proposed explicitly utilizing the explainable interaction of LiDAR and camera modalities for multi-modal fusion-based place recognition. We first proposed the sparse depth supervision and appearance rendering as the specific form of cross-modal interaction and then jointly exploited them in a novel FPR network called EINet. Moreover, we provided a new NUSC-PR benchmark to standardize the training schemes and evaluation protocols in this paper. 
Extensive experiments conducted on the NUSC-PR demonstrated the effectiveness of the proposed explicit interaction for fusion-based place recognition. The efficient inference process also allows us to deploy EINet in real-world applications such as robots and autonomous driving. Both the implementation of our method and benchmark have been released as open-source to facilitate future research.


\bibliographystyle{ieeetr}

\footnotesize{
\bibliography{new}}

\end{document}